\documentclass[11pt,twoside]{article}

\usepackage{palatino, geometry, url}

\usepackage{cite}
\usepackage{amsmath,amssymb,amsfonts}
\usepackage{graphicx}
\usepackage{array}

\allowdisplaybreaks

\usepackage{fancyhdr}
\usepackage{cite}
\usepackage{graphicx}
\usepackage{textcomp}
\usepackage{wrapfig}
\usepackage{amsmath,amsfonts}
\usepackage{algorithmic}
\usepackage{algorithm}
\usepackage{array}
\usepackage[caption=false,font=normalsize,labelfont=sf,textfont=sf]{subfig}
\usepackage{booktabs}    % <-- REQUIRED for \toprule etc.
\usepackage{textcomp}
\usepackage{stfloats}
\usepackage{url}
\usepackage{verbatim}
\usepackage{graphicx}
\usepackage{caption}
\usepackage{xcolor}
\usepackage[justification=justified,singlelinecheck=false]{caption}
\makeatletter
\let\NAT@parse\undefined
\makeatother

\usepackage[hidelinks]{hyperref}

\begin{document}

\title{Multi-Level Temporal Graph Networks with Local-Global Fusion for Industrial Fault Diagnosis}

\author{Bibek Aryal${}^\dag$, Gift Modekwe${}^\dag$, and Qiugang Lu${}^{\dag}$
\thanks{Corresponding author: Qiugang Lu; Email: jay.lu@ttu.edu}\\
\\
{\small ${}^\dag$Department of Chemical Engineering, Texas Tech University, Lubbock, TX 79409, USA}
\\
}
%\date{January 31${}^{st}$, 2023}
\date{}
\maketitle

\begin{abstract}
	
Fault detection and diagnosis are critical for the optimal and safe operation of industrial processes. The correlations among sensors often display non-Euclidean structures where graph neural networks (GNNs) are widely used therein. However, for large-scale systems, local, global, and dynamic relations extensively exist among sensors, and traditional GNNs often overlook such complex and multi-level structures for various problems including the fault diagnosis. To address this issue, we propose a structure-aware multi-level temporal graph network with local-global feature fusion for industrial fault diagnosis. First, a correlation graph is dynamically constructed using Pearson correlation coefficients to capture relationships among process variables. Then, temporal features are extracted through long short-term memory (LSTM)-based encoder, whereas the spatial dependencies among sensors are learned by graph convolution layers.  A multi-level pooling mechanism is used to gradually coarsen and learn meaningful graph structures, to capture higher-level patterns while keeping important fault-related details. Finally, a fusion step is applied to combine both detailed local features and overall global patterns before the final prediction. Experimental evaluations on the Tennessee Eastman process (TEP) demonstrate that the proposed model achieves superior fault diagnosis performance, particularly for complex fault scenarios, outperforming various baseline methods.
	
\end{abstract}


\section{Introduction}

For industrial processes, fault diagnosis (FD) is crucial to ensure their safe operation and compliance with required specifications. It also plays a key role in moving toward more intelligent and automated systems \cite{chen2021graph}. Faults in sensors, actuators, or control loops can result in deviations of variables from desired behaviors, which lead to product quality degradation, significant financial losses, and even safety hazards. Therefore, it is essential for industrial processes to be equipped with intelligent monitoring technologies that can provide timely and reliable fault detection/diagnosis, assisting to reduce economic losses while ensuring safety \cite{kovalenko2024graph}. %Identifying the root cause of a fault often requires considerable time \cite{kovalenko2024graph}. %Variations in equipment parameters can result in defective products, causing production delays and material wastage. These issues ultimately contribute to increased operational costs. Therefore, maintaining equipment in optimal conditions and promptly addressing system faults are crucial for minimizing production expenses \cite{kovalenko2024graph}. 

Fault detection refers to the process of discovering when variables deviate from their normal operating conditions due to faults such as equipment failure, whereas fault diagnosis focuses on identifying different causes/types of faults \cite{ming2017review}. Over the past few decades, a wide range of fault detection and diagnosis (FDD) techniques have been developed. These approaches are generally classified into three main categories: data-driven \cite{qin2009data, lundgren2022data}, model-based \cite{isermann2005model, olivier2009model}, and knowledge-based methods \cite{chi2022knowledge}. Early fault diagnosis methods were primarily model-based, relying on physical models or observers to detect faults through residual analysis. With advancement in sensor technology and data storage, data-driven approaches based on neural networks have gained attention due to their strong data processing capabilities.

Traditional statistical FD methods such as principal component analysis and Fisher discriminant analysis can effectively deal with high-dimensional, noisy, and highly correlated data, by projecting them onto a lower-dimensional subspace that preserves the critical variance characteristics of process behaviors \cite{jiang2013fault, yang2019fault}. However, since these methods are primarily linear dimensionality reduction techniques, this might lead to the loss of critical information and thus poor fault diagnosis performance. As a result, their application is generally restricted to small-scale systems with a limited number of variables and relatively simple correlations\cite{al2023improving}.

Shallow learning methods, such as support vector machines \cite{yin2016recent}, k-nearest neighbor\cite{samanta2016knn}, Gaussian mixture model \cite{yan2017gaussian}, and artificial neural network (ANN)\cite{moosavi2015ann}, have been effectively employed to address fault diagnosis as a classification task, showing promising results in various industrial applications \cite{wu2018deep,song2022multi}. %As a supervised learning technique, SVMs require a large volume of labeled data and struggle with imbalanced datasets, making them costly and often impractical for real-world chemical processes. ANNs, on the other hand, tend to exhibit slow convergence and are susceptible to training instabilities. These shallow learning models possess limited representation capacity, making it challenging to achieve high fault diagnosis accuracy. Furthermore, their reliance on accurately labeled and complete datasets limits their effectiveness, leaving vast quantities of unlabeled data underutilized\cite{song2022multi}.
Recently, deep learning (DL) has emerged as a powerful alternative, consistently outperforming shallow learning methods across various domains, including fault diagnosis. Convolutional neural networks (CNNs) are among the most prominent architectures in deep learning \cite{janssens2016convolutional,xiao2023graph}. %One of the earliest applications of CNNs in fault diagnosis was presented by Janssens et al.\cite{janssens2016convolutional}, who employed CNNs for bearing fault detection. Their approach used the spatial structure of the input data to effectively capture the covariance patterns in the frequency decomposition of accelerometer signals\cite{xiao2023graph}. 
Other deep neural network architectures, such as deep belief networks, recurrent neural networks, and residual networks, have also been widely adopted as representation learners for fault diagnosis \cite{yu2023challenges,wang2020novel,zhu2022application, wen2020transfer}. However, in research domains that involve non-Euclidean structured data, conventional models often show sub-optimal performance due to inherent limitations in handling irregular graph-like structures.

\begin{figure}[tbh]
	\centering
	\includegraphics[width=0.60\textwidth]{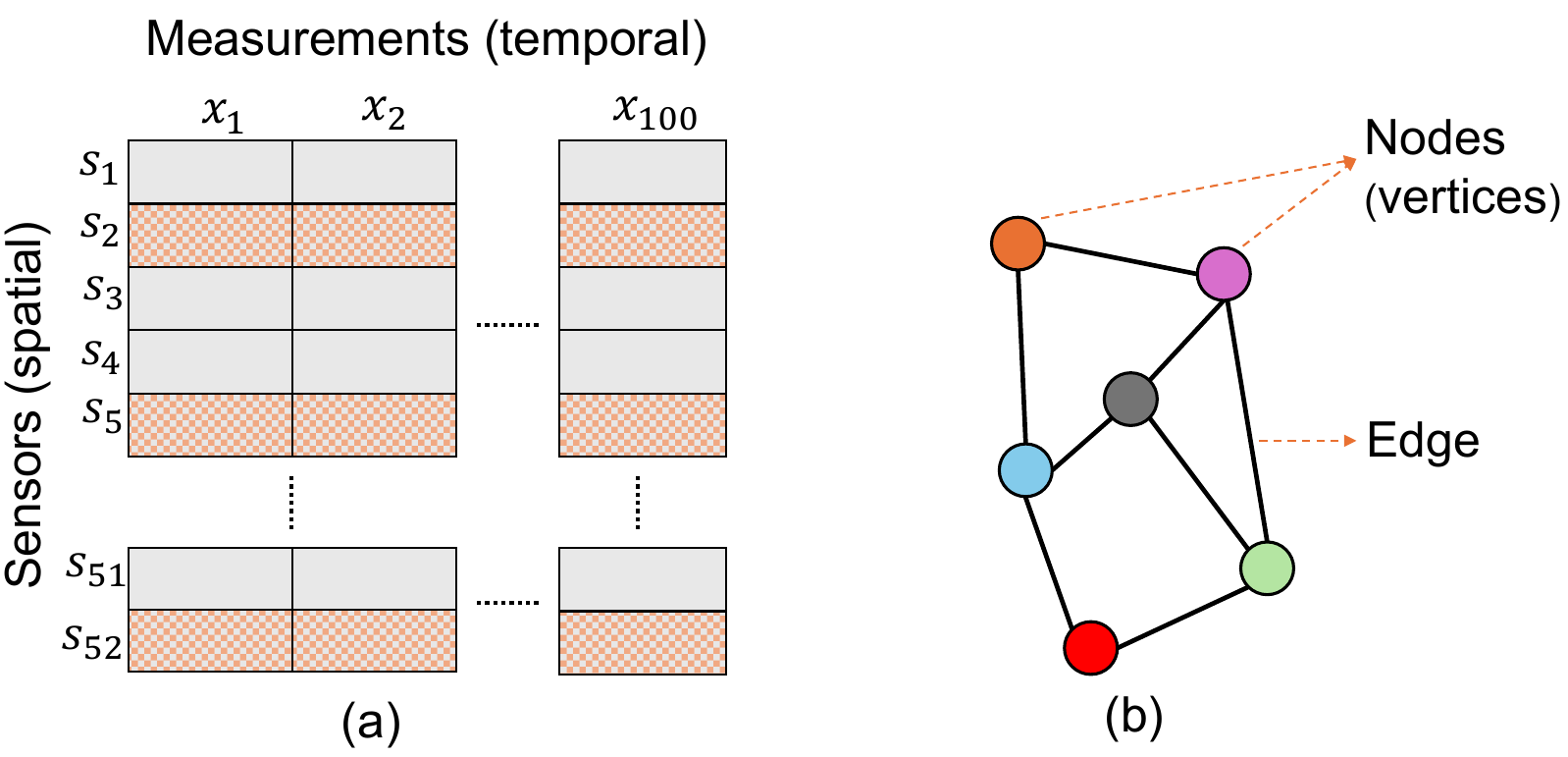}
	\caption{(a): Industrial operating data often display not only localized patterns but also global-level correlations between variables (sensors), e.g., variables $s_2$, $s_5$, $s_{52}$ may be related although they are far-apart spatially;  (b): The representation of a typical graph with circles as nodes and lines as edges.}
	\label{fig0}
\end{figure}

Graph neural network (GNNs) are capable of handling non-Euclidean data structures, making them more suitable for modeling complex relationships in real-world industrial processes\cite{defferrard2016convolutional}. In GNN-based fault diagnosis, variables are often represented as nodes in a graph, with edges modeling the relationships or interactions between them. This structure enables GNNs to effectively capture relationships among process variables, leading to more accurate and robust fault diagnosis \cite{xiao2023graph,liu2024causality,zhao2020multivariate}. Nonetheless, most GNNs rely on message passing between directly connected nodes, limiting their ability to capture \textit{global structural patterns} that are critical for industrial time-series data \cite{al2023improving}. Moreover, most reported GNNs consist of only a single level of granularity and are not suitable for tasks where multi-level structures are critical. Although the DL community has reported Multi-level GNN (ML-GNN) methods, e.g., DIFFPOOL \cite{ying2018hierarchical}, the usage of ML-GNN for fault diagnosis is rather limited \cite{zhang2021hierarchical,chen2023bayesian,zhong2023hierarchical}. The identified knowledge gaps include: (1) existing pooling methods often fail to capture global correlation among variables, which is crucial for complex systems where far-apart variables may still be correlated \cite{jiang2024hierarchical} (see Fig. \ref{fig0} (a)); and (2) the reported GNNs are mostly static and cannot capture multi-level time-varying relations between variables
due to system dynamics\cite{zhong2023hierarchical}. 
To address these gaps, we propose a multi-level temporal GNN model that combines spatio-temporal feature extraction with global context awareness, termed as LGF-MLTG. By combining localized message passing, multi-level graph abstraction, and global feature fusion, the proposed framework captures both detailed local interactions and broader plant-level behaviors. This leads to a more complete representation of process dynamics to enable improved fault diagnosis. The main contributions are summarized as follows:

\begin{itemize}
	\item We propose a multi-level temporal GNN for industrial fault diagnosis that jointly models temporal dynamics and graph-structured variable interactions, to effectively capture both structural and dynamical dependencies in industrial process data.
	
	\item To enhance the spatial expressiveness of node features, we introduce global features (GFs) integration into the ML-GNN framework, which can enable the model to capture long-range correlations between variables. 
	
	\item A comprehensive ablation study is conducted to assess the proposed LGF-MLTG framework on fault diagnosis.
	
\end{itemize}

The rest of the paper is organized as below:  Section II presents the basic concepts used in this work, and Section III describes in detail the proposed H-GNN method. Section IV provides the experimental validation of proposed method, followed with the Conclusion in Section V. 

\section{Preliminaries}
\label{sec: Preliminaries}
%\subsection{CNNs in Fault Diagnosis}
%CNNs are a sort of deep neural network that has been effectively used to solve various classification problems.  CNNs are made up of many filtering and classification steps that extract features automatically.  The filtering step consists of several layers, including the convolution layer, nonlinear activation layer, batch normalization layer, and pooling layer.  The classification stage is a multilayer perceptron made up of many Fully Connected layers\cite{sonmez2023new}.  

\subsection{Graph Neural Networks}
A basic representation of a graph (as in Fig. \ref{fig0} (b)) is $ G = (V, E) $, where $V$ is a set of nodes and $E$ is a set of edges. Let $v_i \in V $ represent a node and $e_{i,j}\in E$ be an edge between $v_i$ and $v_j$. The neighborhood of a node $v_i\in V$ then can be defined as $\mathcal{N}(v_i) = \{v_j \in V | e_{i,j} \in E\}$. The connectivity between nodes is depicted by an adjacency matrix \( \mathbf{A} \in \mathbb{R}^{N \times N} \), where \( N = |\mathcal{V}| \) denotes the number of nodes. Specifically, \( A_{ij} = 1 \) if \( e_{i,j} \in E \) and \( i \neq j \); otherwise, \( A_{ij} = 0 \). In an undirected graph, \( A_{ij} \) represents an edge connection between \( v_i \) and \( v_j \). 

In real-world scenarios, graphs often have node features or attributes, represented by a feature matrix \( \mathbf{X} \in \mathbb{R}^{c \times N} \), where \( c \) denotes the dimensionality of each node's feature vector. The inputs to GNNs contain two components:  the node feature matrix \( \mathbf{X}\) and the adjacency matrix \( \mathbf{A} \). The outputs of GNNs are computed via a sequence of message-passing and aggregation operations, typically expressed below:
\begin{equation}
		\mathbf{Z}_{\text{GNN}} = \text{softmax} \left( \text{GNN}^{(l)} \left( \cdots, \text{GNN}^{(2)} \left( \mathbf{A}, \text{GNN}^{(1)}(\mathbf{A}, \mathbf{X}) \right) \right) \right).
\end{equation}
Here, \( \text{GNN}^{(k)} \) represents the operation performed at the \( k \)-th layer, and \( l \) denotes the total number of layers. The softmax function is applied at the output layer for classification tasks.
%\begin{figure}[tbh]
%	\centering
%	\includegraphics[width=0.2\textwidth]{graph}
%	\caption{Graph representation}
%	\label{fig2}
%\end{figure}
\subsection{GraphSAGE}
\begin{figure}[tbh]
	\centering
	\includegraphics[width=0.60\textwidth]{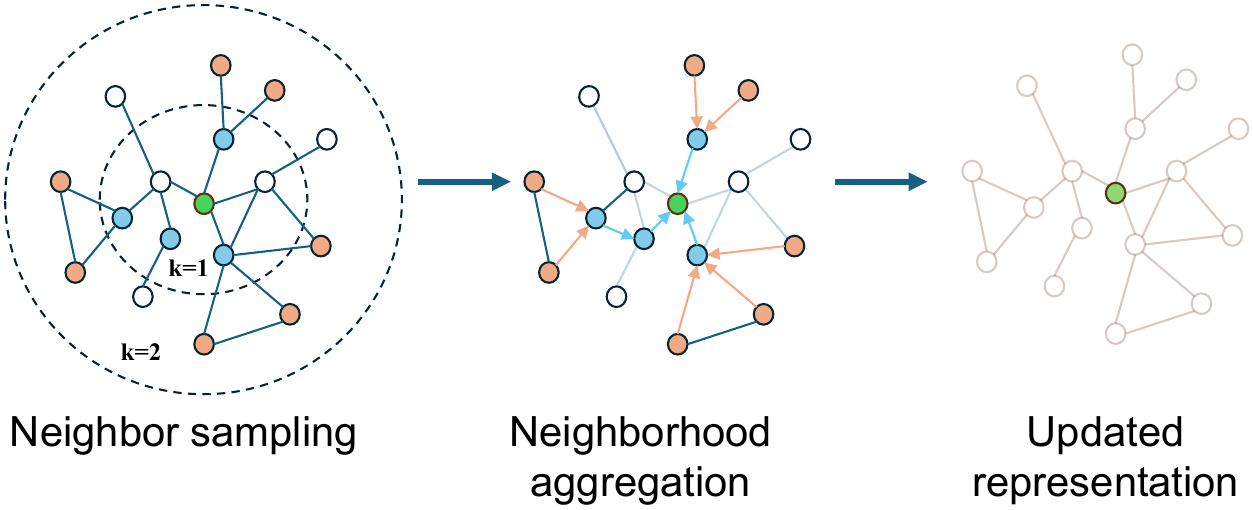}
	\caption{Main components in GraphSAGE: neighbor sampling, feature aggregation, and node representation update.}
	\label{fig3}
\end{figure}

The main idea behind GraphSAGE is to learn node representations by aggregating information from a \textit{sampled set} of neighbors\cite{jiawei2024graphsage}. GraphSAGE is typically structured as a \( K\)-layer architecture, where each layer performs three key operations for a given node \( v \): (1) neighbor sampling, (2) neighborhood aggregation, and (3) representation update. Fig. \ref{fig3} shows how message passing is conducted in GraphSAGE.

At the \( i \)-th layer, the model first samples a fixed-size neighborhood \( \mathcal{N}(v) \) for node \( v \). It then aggregates information from the neighbors according to the following rule:
\begin{equation}
	\mathbf{h}_{\mathcal{N}(v)}^{(i-1)} = \texttt{AGGREGATE}^{(i)}\left( \left\{ \mathbf{h}_u^{(i-1)}, \forall u \in \mathcal{N}(v) \right\} \right)
\end{equation}
Finally, the node representation is updated by combining the aggregated neighborhood information with the node's own previous representation:
\begin{equation}
	\mathbf{h}_v^{(i)} = \sigma \left( \mathbf{W}^{(i)} \texttt{CONCAT} \left( \mathbf{h}_v^{(i-1)}, \mathbf{h}_{\mathcal{N}(v)}^{(i-1)} \right) \right),
\end{equation}
where \( \sigma(\cdot) \) denotes a non-linear activation function (e.g., ReLU), \( \mathbf{W}^{(i)} \) is the learnable weight matrix at the \( i \)-th layer, and \( \texttt{CONCAT}(\cdot) \) represents vector concatenation. The final node representation \( \mathbf{h}_v^{(K)} \) after \( K \) layers encodes both the node’s own features and the structural information from its multi-hop neighborhoods.

%%%%%%%%%%%%%%%%%%%%%%%%%%%%%%%%%%%%%%%%%%

\section{Methodology}
\label{sec:Network}
\begin{figure*}
	\centering
	\includegraphics[width=0.90 \textwidth]{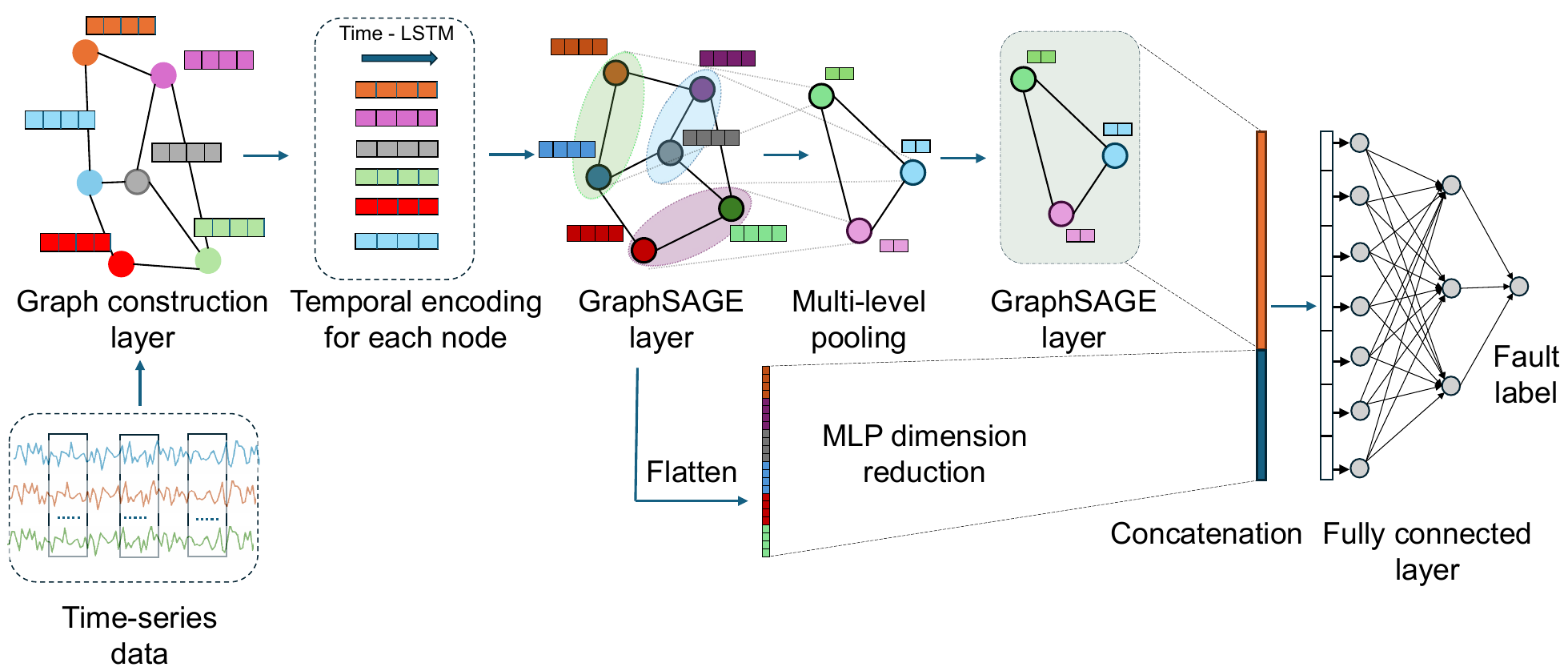}
	\caption{Illustration of the proposed LGF-MLTG framework for fault diagnosis of industrial processes. }
	\label{fig4}
\end{figure*}

This section introduces our proposed LGF-MLTG architecture developed for fault diagnosis of industrial processes. As in Fig. \ref{fig4}, first, graphs are constructed from time-series data and then LSTM network encodes each node's embedding to obtain the temporal features. These updated embeddings are used as an input to GraphSAGE layers to extract spatial correlations, followed by a pooling layer to coarsen the graph. Finally, a feature fusion stage integrates the fine-grained local details with the broader global patterns, before passing through fully-connected layers for fault diagnosis. 
\subsection{Graph-Based Representation of Multivariate Time-Series Data}
\subsubsection{Data Preprocessing}
To ensure uniform scaling across variables and to stabilize model training, $z$-score normalization is applied to each process variable, with the entire training data across all fault classes. With normalized data, to capture local temporal dynamics, we apply sliding windows of length $w$ with stride one to all $N$ variables. Define the moving window at time $t$ as $\mathbf{X}_{t}=[\mathbf{x}_t^1,\ldots,\mathbf{x}_t^N]\in\mathbb{R}^{w\times N}$, where past $w$ values of each variable $i$ are stacked into $\mathbf{x}_{t}^i=x^{i}_{t-w+1:t}\in\mathbb{R}^{w}$. Each segment $\mathbf{X}_{t}$ is used to construct a dynamic graph snapshot that reflects the statistical relationships among variables during that window. Section IV shows details about the dataset. 

\subsubsection{Dynamic Graph Construction via Sliding Window}
As above, each graph snapshot $G_t = (V, \mathbf{A}_t)$ is constructed from a window segment \( \mathbf{X}_{t} \). The node set $V = \{v_1, v_2, \ldots, v_N\}$ corresponds to process variables, and the edges $\mathbf{A}_t$ represent (weighted) pairwise connection between those variables. The strength of the edge between two variables $i$ and $j$ is computed using the Pearson correlation coefficient:
\begin{equation}
	\rho_{ij}^t = \frac{(\mathbf{x}_t^i-\bar{x}^i\cdot \mathbf{1})^\top (\mathbf{x}_t^j-\bar{x}^j\cdot \mathbf{1})}{\|(\mathbf{x}_t^i-\bar{x}^i\cdot \mathbf{1})\|_2\cdot \|(\mathbf{x}_t^j-\bar{x}^j\cdot \mathbf{1})\|_2},
\end{equation}
where $\mathbf{x}^i_t$ denotes the value vector of the \( i \)-th variable in that window, and \( \bar{x}^i \) is the window mean of variable \( i \).

\begin{figure}[!t]
	\centering
	\includegraphics[width=0.60\textwidth]{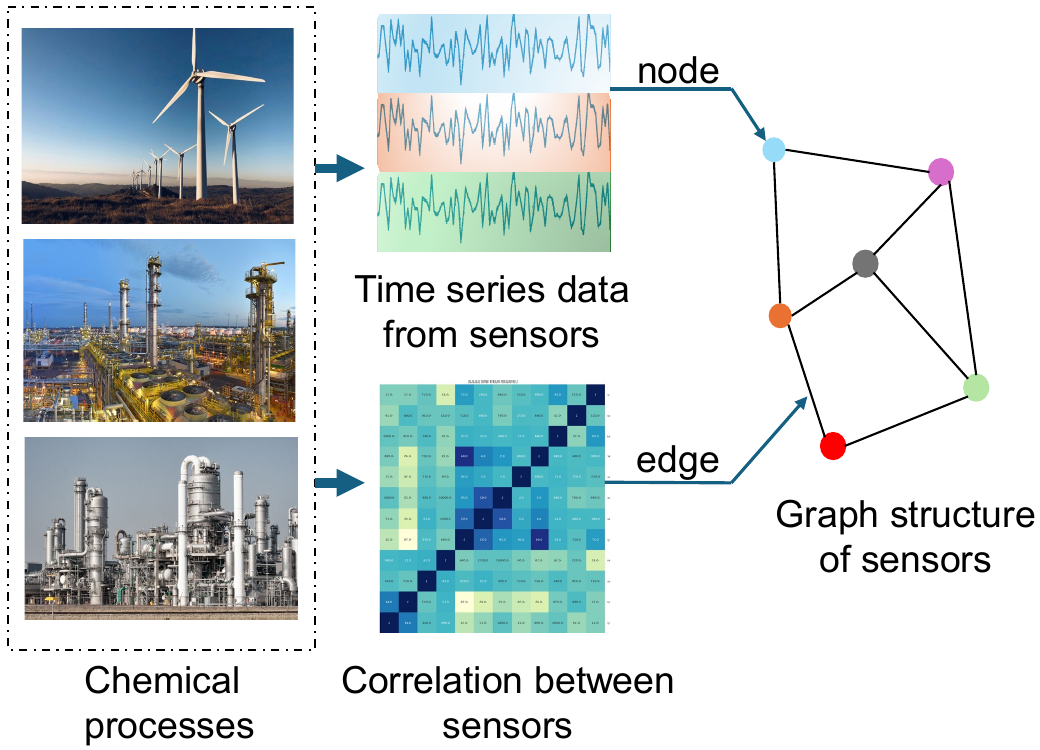}
	\caption{Graph construction from a window of multivariate time-series data where sensors/variables become the nodes and their relations constitute the edges.}
	\label{fig5}
\end{figure}

The resulting correlation matrix \( \mathbf{A}_t \in \mathbb{R}^{N \times N} \) serves as the weighted adjacency matrix for the graph. A hard threshold \( \delta \) is applied to promote sparsity and reduce spurious correlations:

\begin{equation}
	A_{ij}^t =
	\begin{cases}
		\rho_{ij}^t, & \text{if } |\rho_{ij}^t| \geq \delta, \\
		0, & \text{otherwise}. \label{eq: Adjacency}
	\end{cases}
\end{equation}
This approach produces a temporal sequence of graphs \( \{ G_t \}_{t=1}^{M} \), where each graph captures the short-term dependency structure among process variables. Fig. \ref{fig5} shows the process of constructing a graph from multivariate time-series data. These graph snapshots form the foundation for subsequent spatio-temporal modeling.

\subsection{Spatio-Temporal Feature Extraction} 
A two-stage spatio-temporal encoding framework is employed to effectively capture the temporal correlations as well as spatial dependencies within multivariate process data. First, temporal features are extracted from each node's time segment using an unidirectional LSTM network. These features are then propagated across the graph structure using GraphSAGE to model inter-variable relationships.
\subsubsection{Temporal Feature Encoding Using LSTM Networks}

Given a temporal window of length \( w \), the feature of each node \( v_i \in V \) is a univariate time-series segment \( \mathbf{x}^i_t \in \mathbb{R}^{w} \). Each segment is passed through an unidirectional LSTM network to extract temporally-aware node features. Specifically, the feature vector of each node: $\mathbf{x}_t^i$, will be passed through \textit{the same} LSTM network as below: $\forall s\in\{t-m+1,\ldots,t\}$,
\begin{align}
	\mathbf{i}^{(i)}_s &= \sigma(\mathbf{W}_i x_s^{(i)} + \mathbf{U}_i \mathbf{h}_{s-1}^{(i)} + \mathbf{b}_i),  \label{eq: LSTM_begin}\\
	\mathbf{f}_s^{(i)} &= \sigma(\mathbf{W}_f x_s^{(i)} + \mathbf{U}_f \mathbf{h}_{s-1}^{(i)} + \mathbf{b}_f), \\
	\mathbf{o}_s^{(i)} &= \sigma(\mathbf{W}_o x_s^{(i)} + \mathbf{U}_o \mathbf{h}_{s-1}^{(i)} + \mathbf{b}_o), \\
	\mathbf{g}_s^{(i)} &= \tanh(\mathbf{W}_g x_s^{(i)} + \mathbf{U}_g \mathbf{h}_{s-1}^{(i)} + \mathbf{b}_g), \\
	\mathbf{c}_s^{(i)} &= \mathbf{f}_s^{(i)} \odot \mathbf{c}_{s-1}^{(i)} + \mathbf{i}_s^{(i)} \odot \mathbf{g}_s^{(i)}, \\
	\mathbf{h}_s^{(i)} &= \mathbf{o}_s^{(i)} \odot \tanh(\mathbf{c}_s^{(i)}), \label{eq: LSTM_end}
\end{align}
where the superscript $(i)$ denotes the $i$-th variable/node, \( \mathbf{h}_s^{(i)}\) is the hidden state at timestep \(s \), \( \mathbf{c}_s^{(i)}\) is the cell state, and \( \sigma(\cdot) \) denotes the sigmoid activation. The final hidden state \( \mathbf{h}_t^{(i)} \in \mathbb{R}^F \) is used as the updated feature representation for node \( v_i \) that becomes the input to the GraphSAGE model.

\subsubsection{Spatial Dependency Modeling via GraphSAGE}
The temporal features $[\mathbf{h}_t^{(1)},\ldots,\mathbf{h}_t^{(N)}]$ obtained from the LSTM network form the initial node embeddings. These node features are passed through GraphSAGE layers to capture the structural dependencies encoded in the graph \( G_t \). GraphSAGE updates node representations by aggregating information from a node’s local neighborhood. The update rule at layer \(g+1 \) is given by (omitting the time $t$ hereafter): 
\begin{equation}
		\mathbf{h}_v^{(g+1)} = \sigma \left( \mathbf{W}^{(g)} \texttt{CONCAT} \left( \mathbf{h}_v^{(g)}, \texttt{MEAN} \left( \left\{ \mathbf{h}_u^{(g)} : u \in \mathcal{N}(v) \right\} \right) \right) \right). \label{eq: Graph_SAGE_0}
\end{equation}
Here, \( \mathbf{h}_v^{(g)} \) denotes the feature vector of node \( v \) at layer \( g \), \( \mathcal{N}(v) \) denotes its neighbors, \( \sigma(\cdot) \) is a non-linear activation (e.g., ReLU), and \( \mathbf{W}^{(g)} \) are learnable weights. The \texttt{MEAN} aggregator computes the element-wise mean of the neighbor feature vectors. This message-passing framework allows each node to combine both its own temporal dynamics and spatial information from surrounding variables. We define the  feature vector of each node after $G$ GraphSAGE layers as $\mathbf{h}_v^{(G)}\in\mathbb{R}^d$, where $d$ is the feature dimension, $\forall v \in \{1,\ldots,N\}$. The updated graph and its feature matrix form the initial input to the hierarchical pooling layers as below, i.e., $\mathbf{H}^{(0)}:=[\mathbf{h}_{1}^{(G)},\ldots,\mathbf{h}_{N}^{(G)}]^\top\in\mathbb{R}^{N \times d}$.

\subsection{Multi-Level Representation Learning}
A multi-level graph structure is used to improve the model's ability to learn from data \textit{at different levels}. Typically, faults in industrial processes originate within localized units affecting a subset of process variables, and subsequently propagate throughout the system. To address this, the proposed LGF-MLTG framework employs graph pooling to coarsen nodes, effectively treating variable subgroups as ``super-nodes''. This multi-level representation improves fault diagnosis by capturing both local problems and their broader system-level impact, which helps the model to gradually simplify node features while keeping important local connections and overall patterns.
%\begin{figure}[!t]
%	\centering
%	\includegraphics[width=3.0in]{heirarchical_pooling}
%	\caption{Illustration of hierarchical pooling}
%	\label{fig7}
%\end{figure}

\subsubsection{Multi-Level Graph Construction via Node Pooling}
Multi-level graph construction is achieved using a graph pooling mechanism that coarsens the input graph by clustering nodes into super-nodes without requiring any domain knowledge. The pooling layers learns a cluster \textit{assignment matrix} over the nodes.  At layer $l$, node embeddings generated by the GNN at layer $l-1$ are used to guide pooling. In this way, GNNs not only extracts features but also generates embeddings that support hierarchical pooling. For the $l$-th pooling layer, define the feature matrix of the input graph with $n_l$ nodes as \( \mathbf{H}^{(l)} \in \mathbb{R}^{n_l \times d}\), where $n_0=N$ and $n_l$ is the supernode number at layer $l$, and adjacency matrix as \( \mathbf{A}^{(l)} \in \mathbb{R}^{n_l \times n_l} \). A soft clustering assignment matrix \( \mathbf{S}^{(l)} \in \mathbb{R}^{n_l \times n_{l+1}} \) can be learned via a GNN layer \cite{ying2018hierarchical}
\begin{equation}
	\mathbf{S}^{(l)}=\text{softmax}\left(\text{GNN}^{(l)}(\mathbf{A}^{(l)},\mathbf{H}^{(l)})\right),  \label{eq:assignment_matrix}
\end{equation}
where $n_{l+1}$ is the number of clusters (super-nodes) in the next layer. Each row of $\mathbf{S}^{(l)}$ specifies the likelihood of assigning a node to each super-node.  Then, the next-layer pooled node features and adjacency matrix are computed as:
\begin{align}
	\mathbf{H}^{(l+1)} &= \mathbf{S}^{(l)\top} \mathbf{H}^{(l)} \in \mathbb{R}^{n_{l+1} \times d}, \label{eq:node_features} \\
	\mathbf{A}^{(l+1)} &= \mathbf{S}^{(l)\top} \mathbf{A}^{(l)} \mathbf{S}^{(l)} \in \mathbb{R}^{n_{l+1} \times n_{l+1}}. \label{eq:adjacency}
\end{align}
This process reduces the graph size while maintaining its structural integrity. Fig. \ref{fig4} (top) shows the pooling mechanism where local nodes are pooled into super-nodes represented by distinct colors. The resulting coarsened graph captures higher-level dependencies among groups of variables, which may correspond to functional sub-systems or tightly coupled process units. Finally, the updated feature matrix and adjacency matrix are then fed into another GNN (or GraphSAGE) layer to study the interactions between super nodes.

\subsubsection{Fusion of Local and Global Node Representation}
To preserve both detailed local information, higher-level, and abstract global-level patterns, features from the original node space and the coarsened graph are concatenated to form the final representation (see Fig. \ref{fig4}). Let $\mathbf{H}^{(L)}\in\mathbb{R}^{n_L\times d}$ denote the feature matrix after the last GraphSAGE layer (shaded light green rectangle in Fig. \ref{fig4}) where $n_L$ is the number of super-nodes. With the initial feature matrix $\mathbf{H}^{(0)}\in\mathbb{R}^{N\times d}$, the local-global fused representation is given by:
\begin{eqnarray}
	&\tilde{\mathbf{H}}^{(0)} = \texttt{MLP}(\mathbf{H}^{(0)}), \label{eq: global}\\
	&\hat{y}_t = \texttt{MLP} \left( \texttt{CONCAT} \left( \texttt{Vec}(\mathbf{H}^{(L)}), \texttt{Vec}(\tilde{\mathbf{H}}^{(0)}) \right) \right), \label{eq: final_predict}
\end{eqnarray}
where $\texttt{MLP}(\cdot)$ denotes a multi-layer perceptron, $\texttt{Vec}(\cdot)$ is the vectorization operation, and $\hat{y}_t$ is the predicted fault class label for moving window at time $t$. This hierarchical integration enables the model to leverage both fine-grained patterns, coarse-grained contextual signals, and global-level features, which lead to more robust and interpretable fault diagnosis.

\begin{algorithm}[h]
	\caption{Training Procedure of LGF-MLTG}
	\label{alg:LGF-MLTG}
	\begin{algorithmic}[1]
		\REQUIRE Time-series data $\mathbf{X}^{0}\in \mathbb{R}^{Z\times N}$ of all faults ($N$ variables/nodes, $Z$ samples), learning rate $\eta$
		\ENSURE Trained model parameters $\Theta$
		\STATE Create moving windows of length $w$ from $\mathbf{X}^{0}$; Define current window as $\mathbf{X}_t\in\mathbb{R}^{w\times N}$ with fault label $y_t$ (according to which faulty data $\mathbf{X}_t$ comes from);
		\STATE Compute adjacency matrix $\mathbf{A}_t$ with \eqref{eq: Adjacency};
		\STATE Extract temporal embeddings $\tilde{\mathbf{h}}_t:=[\mathbf{h}_t^{(1)},\ldots,\mathbf{h}_t^{(N)}]$ with LSTM network \eqref{eq: LSTM_begin}-\eqref{eq: LSTM_end} (omit time index $t$ hereafter);
		\STATE Learn spatial features by passing $\{\tilde{\mathbf{h}}_t, \mathbf{A}_t\}$ through $G$ GraphSAGE layers \eqref{eq: Graph_SAGE_0}, and obtain spatiotemporal features $\mathbf{H}^{(0)}:=[\mathbf{h}^{(G)}_0,\ldots,\mathbf{h}^{(G)}_N]^{\top}\in\mathbb{R}^{N \times d}$;
		\STATE Pass $\mathbf{H}^{(0)}$ into MLP to extract \textit{global} spatiotemporal features $\tilde{\mathbf{H}}^{(0)}$ according to \eqref{eq: global};
		\STATE In parallel, pass $\mathbf{H}^{(0)}$ into pooling layers to obtain coarsened graphs $(\mathbf{H}^{(l)}, \mathbf{A}^{(l)})$ with  \eqref{eq:assignment_matrix}-\eqref{eq:adjacency}.
		\STATE Input the last coarsened graph through GraphSAGE layer to obtain $\mathbf{H}^{(L)}$;
		\STATE Fuse global features $\tilde{\mathbf{H}}^{(0)}$ (from Step 5) with \textit{local} coarsened features $\mathbf{H}^{(L)}$ to predict fault label $\hat{y}_t$ with \eqref{eq: final_predict};
		\FOR{$\texttt{epoch} = 1$ to \texttt{epoch number}}
		\STATE Compute the loss $L = \mathcal{L}_{Cross Entropy}(y_t,\hat{y}_t) + \alpha \, \mathcal{L}_{Pool}(\cdot)$, where the pooling loss (with weight $\alpha$) is defined similarly to the loss of DIFFPOOL as in  \cite{ying2018hierarchical};
		\STATE Update $\Theta \leftarrow \Theta - \eta \nabla_\Theta L$;
		\ENDFOR
		\RETURN $\Theta$
	\end{algorithmic}
\end{algorithm}

\section{Case Studies}
\label{sec: simulation}
In this section, the benchmark Tennessee Eastman process (TEP) dataset \cite{rieth2017additional} is used to assess the performance of the proposed LGF-MLTG model.
\subsection{Experimental Validation}
\begin{figure}[!t]
	\centering
	\includegraphics[width=0.90\textwidth]{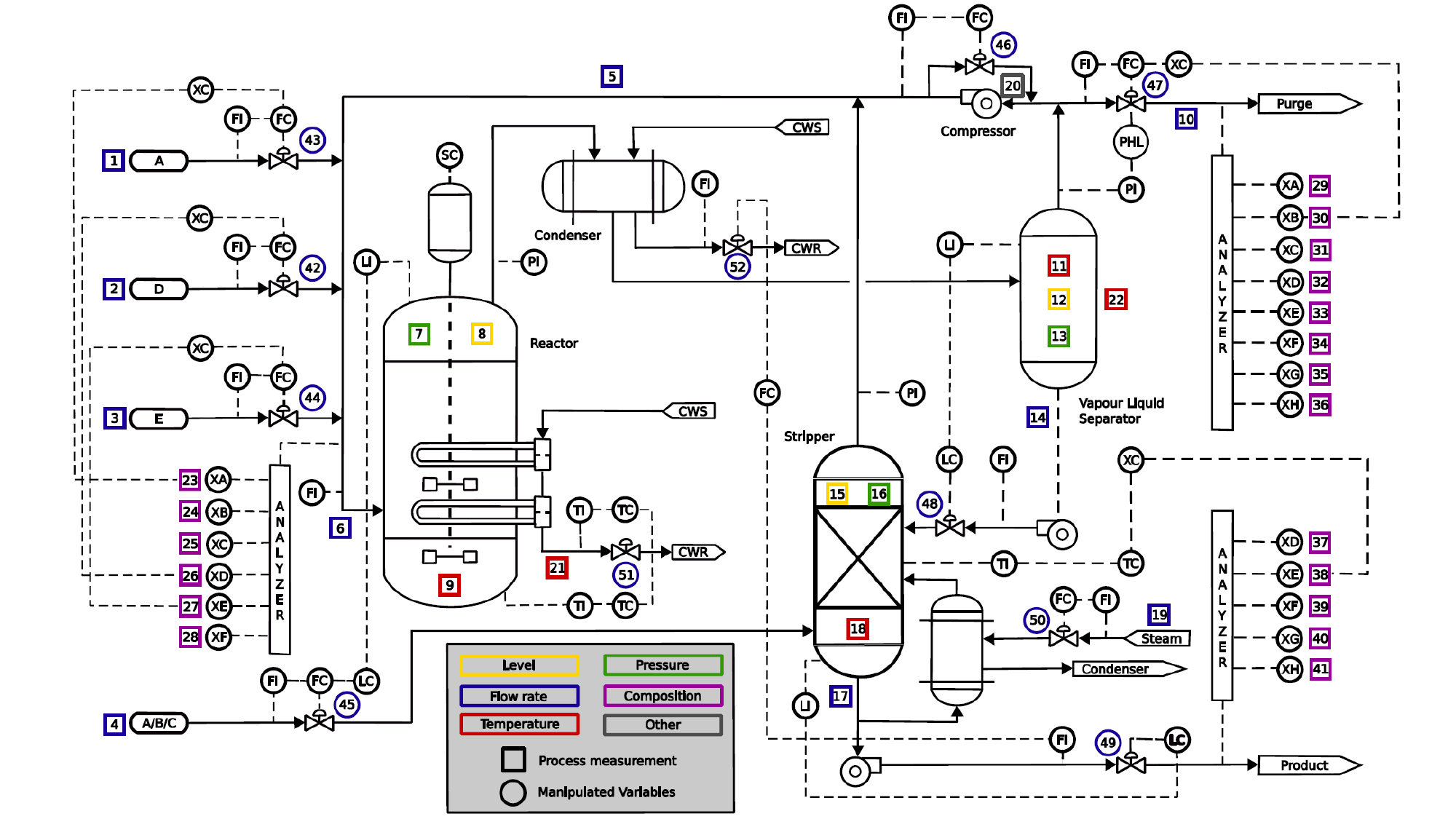}
	\caption{The schematics of the TEP \cite{kovalenko2024graph}.}
	\label{fig9}
\end{figure}

TEP has become a popular benchmark in the community to assess control and fault diagnosis methods \cite{downs1993plant}. It simulates a real industrial process with five units: a reactor, a condenser, a separator, a stripper, and a compressor (see Fig. \ref{fig9}). It consists of 53 process variables, among which 12 are manipulated variables such as speed, flow rates, and valves position, with the other 41 showing variables like pressure, liquid level, and temperature. The given dataset contains 52 variables, where the agitator speed is excluded.  The dataset contains simulations of 20 different faults for each simulation runs. There are 500 simulation runs for training and testing data, respectively. For this study, 110 simulation runs are used (100 for training and 10 for testing). For each fault type, the training and testing datasets consist of 500 and 960 samples, respectively, with process variables recorded every three minutes. In each simulation run, the fault is introduced after 1 hour in the training data and after 8 hours in the testing data. Therefore, the training set includes 480 faulty samples per fault type, while the testing set contains 800 faulty samples.

\begin{figure}[!t]
	\centering
	\includegraphics[width=0.60\textwidth]{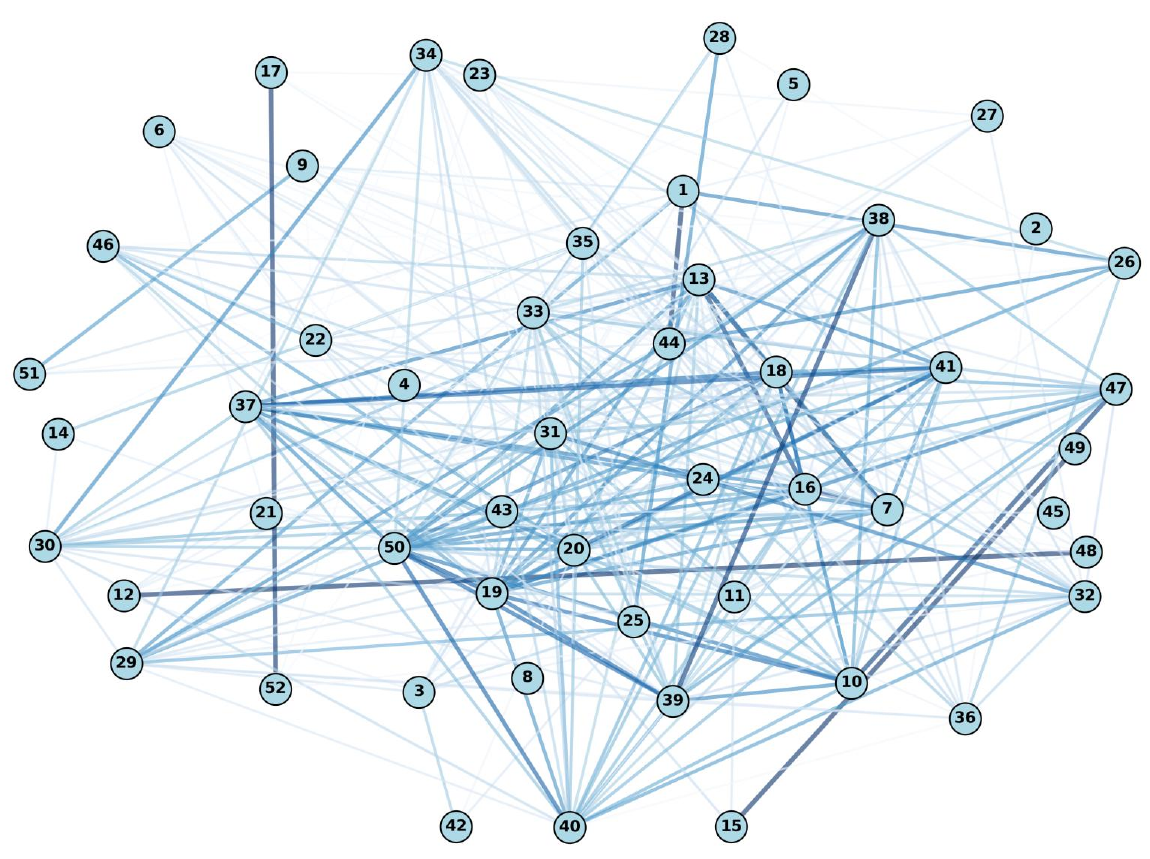}
	\caption{The graph constructed from a moving window, where thick edges indicate large weights (stronger connection).}
	\label{fig8}
\end{figure}
For each faulty dataset, process variables are first preprocessed with $z$-score normalization. Then, a sliding window extracts samples after the standardization. The length of the sliding window is set to be 100 samples with a stride of 1. The successive timesteps of samples in each moving window are to capture temporal correlation between variables. Each window has a label based on which faulty data it comes from. Then adjacency matrices are constructed for each window separately using Pearson's correlation coefficient. Fig.  \ref{fig8} shows one graph constructed for a specific moving window where the thick and thin edges represent strong and weak correlations between variables. In our studies, the threshold of Pearson's correlation coefficient to construct a coarse adjacency matrix is set as $\delta=0.5$. Hence, each window will be represented as weighted graph for the proposed method. The GNN model used for multi-level modeling is built on top of the GraphSAGE architecture, since it is found to have superior performance compared with other GNN models such as GCN \cite{ying2018hierarchical}. The ``mean" variant of GraphSAGE is used, while hierarchical pooling is performed after two GraphSAGE layers, then after the pooling, another graphSAGE layer follows.

All the models are trained on an NVIDIA RTX 4080 GPU to speed up the computation. Adam optimizer is used with a learning rate of 0.003 and a batch size of 32 to balance memory usage and training speed. The loss function employed for the study consists of the cross-entropy loss $\mathcal{L}_{CrossEntropy}(\cdot)$, along with auxiliary loss such as the pooling loss $\mathcal{L}_{Pool}(\cdot)$ (cf. the pseudo-code in Algorithm 1).  The number of soft clusters (supernodes) for pooling layer is set to be 8.

\subsection{Comparison with Baseline Models}
In this section, we will thoroughly compare the proposed LGF-MLTG model against several state-of-the-art DL methods to show the model's effectiveness in capturing hierarchical spatiotemporal correlations of process variables for fault diagnosis. The assessment involves three standard metrics: fault detection rate (FDR), precision, and the F1-score.
\begin{figure}[!t]
	\centering
	\includegraphics[width=0.75\textwidth]{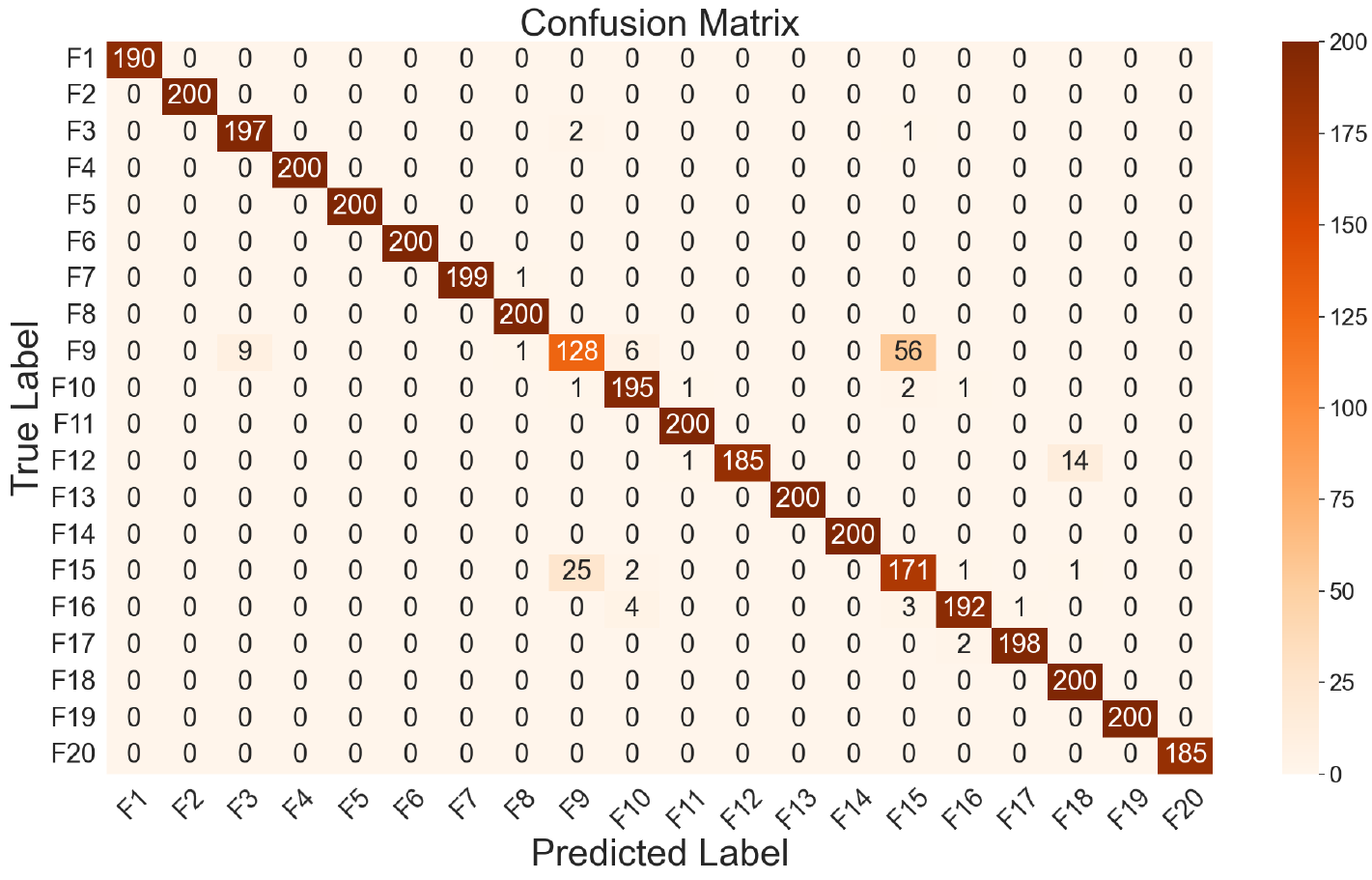}
	\caption{Confusion matrix of the fault diagnosis with our LGF-MLTG model for the test dataset of TEP.}
	\label{fig10}
\end{figure}
The FDR, also referred to as recall, quantifies the model’s ability to correctly identify fault types, defined as:
\begin{equation}
	FDR = \dfrac{TP}{TP + FN}, \label{eq: FDR}
\end{equation}
with the numbers of true positives $TP$ and false negatives $FN$. Similarly, the precision is defined as the ratio of true positives to the total number of positive predictions made by the model for a given fault class:
\begin{equation}
	\text{Precision} = \frac{TP}{TP + FP},
\end{equation}
where \( FP \) represents the number of false positives. This metric reflects the model's ability to minimize false alarms, with higher values indicating better precision. In contrast, the F1-score provides a balanced measure by incorporating both precision and FDR, offering a more comprehensive view of the model’s performance in imbalanced classification settings.
The F1-score is defined as:

\begin{equation}
	F_1 = \frac{2 \times \text{FDR} \times \text{Precision}}{\text{FDR} + \text{Precision}}.
\end{equation}
A higher F1-score indicates a model that performs well on both detection sensitivity and precision.

The models used for comparison include a traditional ANN model, a modified global feature CNN (GF-CNN) introduced in \cite{lu2022enhanced}, and a model combining both LSTM and CNN (LSTM-CNN) introduced in \cite{huang2022novel}. In addition, two representative graph-based baselines are considered: the spatio-temporal GCN (ST-GCN) \cite{yu2018spatio}, which jointly models spatial and temporal dependencies, and the graph attention network (GAT) \cite{velivckovic2018graph}, which employs attention mechanisms to learn adaptive node interactions. %From Table \ref{tab1}, it can be seen that our LGF-MLTG model clearly outperforms the others, especially for Faults 3 and 15 that are known to be difficult to diagnose.
\begin{figure}[!t]
	\centering
	\includegraphics[width=0.60\textwidth]{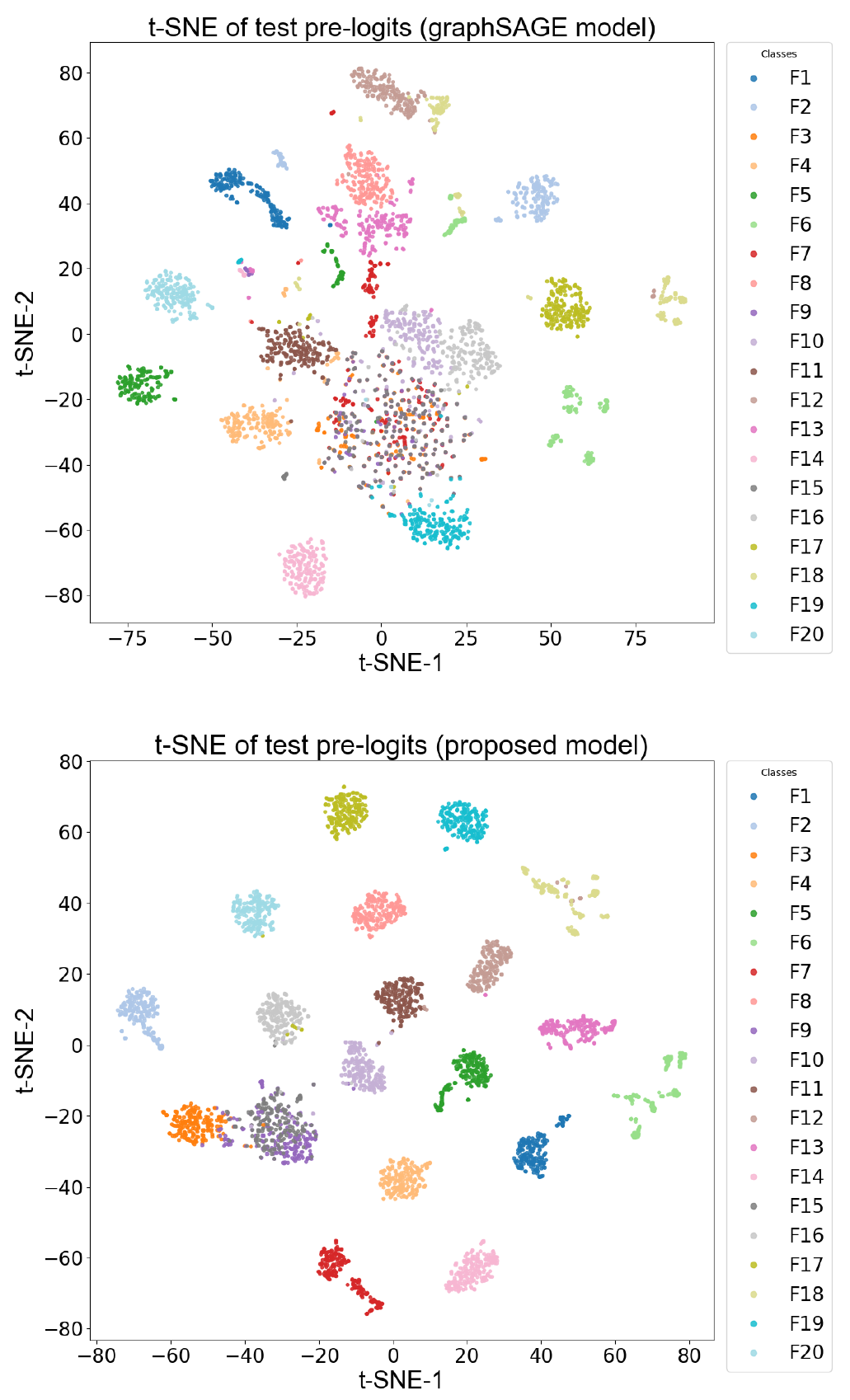}
	\caption{t-SNE plot of testing data in GraphSAGE model (top) and our LGF-MLTG model (bottom).}
	\label{fig12}
\end{figure}
\begin{table*}[ht]
	\tiny
	\centering
	\caption{Performance comparison of LGF-MLTG with baseline models on the TEP dataset}
	\label{tab1}
	\resizebox{\textwidth}{!}{
		\begin{tabular}{c|ccc|ccc|ccc|ccc|ccc|ccc}
			\toprule
			\textbf{Fault} 
			& \multicolumn{3}{c|}{\textbf{LGF-MLTG}} 
			& \multicolumn{3}{c|}{\textbf{LSTM-CNN}} 
			& \multicolumn{3}{c|}{\textbf{GF-CNN}} 
			& \multicolumn{3}{c|}{\textbf{ST-GCN}} 
			& \multicolumn{3}{c|}{\textbf{GAT}} 
			& \multicolumn{3}{c}{\textbf{ANN}} \\
			\cmidrule{2-19}
			& FDR & P & F1 
			& FDR & P & F1 
			& FDR & P & F1 
			& FDR & P & F1 
			& FDR & P & F1 
			& FDR & P & F1 \\
			\midrule
			1  & 100 & 100 & 100 & 100 & 100 & 100 & 100 & 100 & 100 & 100 & 99.5 & 99.7 & 98.9 & 100 & 99.5 & 100 & 100 & 100 \\
			2  & 100 & 100 & 100 & 100 & 100 & 100 & 100 & 100 & 100 & 100 & 100 & 100 & 100 & 100 & 100 & 100 & 100 & 100 \\
			3  & \textbf{\underline{98.5}} & \textbf{\underline{95.6}} & \textbf{\underline{97.0}} & 83.1 & 89.5 & 86.2 & 24.0 & 100 & 38.7 & 50.0 & 93.5 & 65.1 & 66.0 & 36.5 & 47.0 & 90.0 & 88.2 & 89.1 \\
			4  & 100 & 100 & 100 & 100 & 100 & 100 & 100 & 99.0 & 99.5 & 100 & 98.5 & 99.2 & 99.0 & 66.9 & 79.8 & 100 & 96.6 & 98.3 \\
			5  & 100 & 100 & 100 & 100 & 100 & 100 & 100 & 100 & 100 & 98.5 & 97.5 & 98.0 & 98.5 & 97.5 & 98.0 & 100 & 98.0 & 99.0 \\
			6  & 100 & 100 & 100 & 100 & 100 & 100 & 100 & 100 & 100 & 100 & 100 & 100 & 100 & 94.3 & 97.1 & 100 & 100 & 100 \\
			7  & 99.5 & 100 & 99.7 & 100 & 100 & 100 & 100 & 100 & 100 & 100 & 99.5 & 99.7 & 90.5 & 100 & 95.0 & 100 & 99.5 & 99.8 \\
			8  & 100 & 99.0 & 99.5 & 100 & 99.0 & 99.5 & 100 & 100 & 100 & 98.5 & 98.0 & 98.2 & 91.5 & 77.9 & 84.1 & 99.0 & 99.5 & 99.2 \\
			9  & 64.0 & \textbf{\underline{82.1}} & 71.9 & 55.4 & 50.2 & 52.7 & \textbf{\underline{96.5}} & 60.1 & \textbf{\underline{74.0}} & 64.0 & 57.9 & 60.8 & 6.0 & 75.0 & 11.1 & 44.5 & 43.0 & 43.7 \\
			10 & 97.5 & 94.2 & 95.8 & 96.4 & 97.4 & 96.9 & 99.0 & 99.0 & 99.0 & 91.5 & 87.1 & 89.3 & 23.0 & 71.9 & 34.9 & 77.5 & 87.1 & 82.0 \\
			11 & 100 & 99.0 & 99.5 & 99.5 & 100 & 99.7 & 100 & 100 & 100 & 98.5 & 100 & 99.2 & 98.0 & 86.7 & 92.0 & 87.0 & 96.1 & 91.3 \\
			12 & 92.5 & 100 & 96.1 & 93.3 & 100 & 96.6 & 94.0 & 100 & 96.9 & 95.0 & 97.1 & 96.0 & 84.5 & 97.1 & 90.4 & 91.0 & 100 & 95.3 \\
			13 & 100 & 100 & 100 & 99.5 & 100 & 99.7 & 99.5 & 100 & 99.7 & 97.5 & 99.5 & 98.5 & 81.5 & 93.7 & 87.2 & 99.0 & 100 & 99.5 \\
			14 & 100 & 100 & 100 & 100 & 100 & 100 & 100 & 100 & 100 & 100 & 99.5 & 99.7 & 100 & 100 & 100 & 100 & 99.5 & 99.8 \\
			15 & \textbf{\underline{85.5}} & \textbf{\underline{73.4}} & \textbf{\underline{79.0}} & 59.5 & 58.9 & 59.2 & 44.5 & 38.7 & 41.4 & 83.5 & 61.4 & 70.8 & 24.5 & 34.8 & 28.7 & 56.0 & 43.9 & 49.2 \\
			16 & 96.0 & 98.0 & 97.0 & 95.4 & 97.9 & 96.6 & 98.5 & 98.0 & 98.3 & 92.0 & 95.8 & 93.9 & 79.5 & 50.8 & 62.0 & 80.0 & 78.4 & 79.2 \\
			17 & 99.0 & 99.5 & 99.2 & 100 & 100 & 100 & 98.5 & 99.5 & 99.0 & 98.0 & 100 & 99.0 & 96.5 & 99.5 & 98.0 & 98.0 & 99.5 & 98.7 \\
			18 & 100 & 93.0 & 96.4 & 100 & 93.8 & 96.8 & 100 & 93.9 & 96.9 & 100 & 94.0 & 96.9 & 92.5 & 93.9 & 93.2 & 100 & 93.9 & 96.9 \\
			19 & 100 & 100 & 100 & 99.5 & 100 & 99.7 & 100 & 99.5 & 99.8 & 100 & 100 & 100 & 90.0 & 96.3 & 93.0 & 64.0 & 76.2 & 69.6 \\
			20 & 100 & 100 & 100 & 100 & 100 & 100 & 99.5 & 100 & 99.7 & 95.1 & 82.6 & 88.4 & 95.1 & 82.6 & 88.4 & 97.3 & 96.8 & 97.0 \\
			\midrule
			\textbf{Average (\%)} 
			& \textbf{96.6} & \textbf{96.7} & \textbf{96.6}
			& 94.1 & 94.3 & 94.2
			& 92.7 & 94.4 & 92.1
			& 93.4 & 94.2 & 93.4
			& 80.8 & 82.8 & 79.0
			& 81.2 & 82.9 & 81.8 \\
			\bottomrule
		\end{tabular}
	}
\end{table*}

Table~\ref{tab1} presents the comparative performance of the proposed LGF-MLTG model against the five baseline models as above, across 20 fault classes from the TEP dataset. Overall, the LGF-MLTG model outperforms baseline models across all metrics, achieving an average FDR and F1-score of 96.6\%, and precision of 96.7\%. Among the deep learning baselines, LSTM-CNN attains the best performance with an average FDR of  94.1\%, while GF-CNN and ST-GCN show slightly lower accuracy due to their limited ability to capture multi-level process structures. Furthermore, attention-based GAT model shows weaker performance, particularly on difficult fault classes, resulting in a substantially lower average FDR of 80.8\%. Also, the baseline ANN model has a lower F1-score of only 81.8\%. These results clearly indicate that the integration of multi-level pooling, temporal modeling via LSTM, and global feature extraction in LGF-MLTG significantly improves fault classification performance.

\subsection{Ablation Studies}

In this section, ablation experiments are performed to evaluate the contribution of the multi-level graph structure and analyze the influence of incorporating global features for fault diagnosis. To reduce the impact of randomness, each experiment is repeated five times with different random seeds, and the averaged performance is reported.

To evaluate the contribution of individual components in our proposed architecture, we compare five model variants: (1) GraphSAGE, (2) GraphSAGE with LSTM, (3) LGF-MLTG without global features (GF), (4) LGF-MLTG without LSTM, and (5) the full  LGF-MLTG model. The performance is assessed across 20 fault classes with the previous three metrics. Table~\ref{tab:fdr_score} summarizes the results, where the last row shows the averaged metrics across all fault classes. The proposed model achieves the highest performance in all three metrics with an averaged FDR of 96.6\%, F1-score of 96.6\%, and precision of 96.7\%. These results confirm the complementary nature of these components. Introducing LSTM layers improves temporal feature extraction, leading to significant performance gains over baseline models. Similarly, replacing GraphSAGE with hierarchical GNN without global features improves structural learning through hierarchical aggregation, as shown by its F1-score of 95.2\% against 94.1\% for GraphSAGE. Further, adding global features alone (LGF-MLTG without LSTM) shows moderate improvements, but its full potential is realized when combined with LSTM (LGF-MLTG).

\begin{table*}
	\tiny
	\centering
	\caption{Fault diagnosis results for the TEP with different composites for the hierarchical graph network architecture.}
	\label{tab:fdr_score}
	\resizebox{\textwidth}{!}{
		\begin{tabular}{c|ccc|ccc|ccc|ccc|ccc}
			\toprule
			\textbf{Class (Fault ID)} & \multicolumn{3}{c|}{\textbf{GraphSAGE}} & \multicolumn{3}{c|}{\textbf{GraphSAGE with LSTM}} & \multicolumn{3}{c|}{\textbf{LGF-MLTG without GF}} & \multicolumn{3}{c|}{\textbf{LGF-MLTG without LSTM}} & \multicolumn{3}{c}{\textbf{LGF-MLTG}} \\
			\cmidrule{2-16}
			& FDR & Precision & F1 & FDR & Precision & F1 & FDR & Precision & F1 & FDR & Precision & F1 & FDR & Precision & F1 \\
			\midrule
			1  & 100.0 & 99.0 & 99.5 & 100.0 & 100.0 & 100.0 & 100.0 & 100.0 & 100.0 & 100.0 & 100.0 & 100.0 & 100.0 & 100.0 & 100.0 \\
			2  & 99.0 & 100.0 & 99.5 & 100.0 & 100.0 & 100.0 & 100.0 & 100.0 & 100.0 & 100.0 & 100.0 & 100.0 & 100.0 & 100.0 & 100.0 \\
			3  & 94.5 & 72.4 & 82.0 & 98.0 & 71.5 & 82.7 & 94.5 & 81.1 & 87.3 & 93.5 & 92.1 & 92.8 & \textbf{\underline{98.5}} & \textbf{\underline{95.6}} & \textbf{\underline{97.0}} \\
			4  & 100.0 & 99.0 & 99.5 & 100.0 & 100.0 & 100.0 & 100.0 & 100.0 & 100.0 & 100.0 & 99.5 & 99.8 & 100.0 & 100.0 & 100.0 \\
			5  & 100.0 & 99.5 & 99.8 & 100.0 & 100.0 & 100.0 & 100.0 & 99.5 & 99.8 & 100.0 & 100.0 & 100.0 & 100.0 & 100.0 & 100.0 \\
			6  & 100.0 & 100.0 & 100.0 & 100.0 & 100.0 & 100.0 & 100.0 & 100.0 & 100.0 & 100.0 & 100.0 & 100.0 & 100.0 & 100.0 & 100.0 \\
			7  & 100.0 & 99.0 & 99.5 & 100.0 & 93.0 & 96.4 & 100.0 & 100.0 & 100.0 & 100.0 & 100.0 & 100.0 & 99.5 & 100.0 & 99.7 \\
			8  & 99.5 & 96.1 & 97.8 & 99.0 & 100.0 & 99.5 & 100.0 & 98.5 & 99.3 & 100.0 & 100.0 & 100.0 & 100.0 & 99.0 & 99.5 \\
			9  & 54.0 & 58.4 & 56.1 & 50.5 & 63.9 & 56.4 & 52.5 & 67.7 & 59.2 & 45.5 & 62.8 & 52.8 & \textbf{\underline{64.0}} & \textbf{\underline{82.1}} & \textbf{\underline{71.9}} \\
			10 & 82.5 & 97.6 & 89.4 & 98.0 & 97.5 & 97.8 & 96.0 & 100.0 & 98.0 & 92.0 & 87.2 & 89.5 & 97.5 & 94.2 & 95.8 \\
			11 & 99.5 & 99.0 & 99.3 & 100.0 & 99.5 & 99.8 & 99.5 & 100.0 & 99.7 & 99.5 & 99.5 & 99.5 & 100.0 & 99.0 & 99.5 \\
			12 & 93.5 & 99.5 & 96.4 & 94.0 & 100.0 & 96.9 & 95.5 & 100.0 & 97.7 & 93.0 & 99.5 & 96.1 & 92.5 & 100.0 & 96.1 \\
			13 & 93.5 & 100.0 & 96.6 & 98.0 & 99.0 & 98.5 & 98.0 & 100.0 & 99.0 & 100.0 & 98.5 & 99.3 & 100.0 & 100.0 & 100.0 \\
			14 & 100.0 & 100.0 & 100.0 & 100.0 & 99.5 & 99.8 & 100.0 & 100.0 & 100.0 & 99.5 & 100.0 & 99.7 & 100.0 & 100.0 & 100.0 \\
			15 & 56.0 & 67.1 & 61.0 & 59.0 & 69.4 & 63.8 & 72.0 & 67.3 & 69.6 & 68.0 & 58.1 & 62.7 & \textbf{\underline{85.5}} & \textbf{\underline{73.4}} & \textbf{\underline{79.0}} \\
			16 & 98.0 & 92.5 & 95.1 & 99.5 & 99.0 & 99.3 & 100.0 & 95.7 & 97.8 & 92.5 & 95.4 & 93.9 & 96.0 & 98.0 & 97.0 \\
			17 & 99.0 & 100.0 & 99.5 & 98.0 & 100.0 & 99.0 & 99.0 & 100.0 & 99.5 & 99.0 & 98.0 & 98.5 & 99.0 & 99.5 & 99.2 \\
			18 & 99.5 & 93.4 & 96.4 & 93.0 & 93.0 & 93.0 & 100.0 & 95.7 & 97.8 & 100.0 & 93.5 & 96.6 & 100.0 & 93.0 & 96.4 \\
			19 & 100.0 & 96.2 & 98.0 & 100.0 & 100.0 & 100.0 & 100.0 & 100.0 & 100.0 & 99.0 & 95.2 & 97.1 & 100.0 & 100.0 & 100.0 \\
			20 & 97.8 & 98.4 & 98.1 & 99.5 & 100.0 & 99.7 & 100.0 & 100.0 & 100.0 & 97.8 & 98.4 & 98.1 & 100.0 & 100.0 & 100.0 \\
			\midrule
			\textbf{Average (\%)} & 91.1 & 93.6 & 92.3 & 94.3 & 94.3 & 94.1 & 95.3 & 95.3 & 95.2 & 94.0 & 93.9 & 93.8 & \textbf{\underline{96.6}} & \textbf{\underline{96.7}} & \textbf{\underline{96.6}} \\
			\bottomrule
		\end{tabular}
	}
\end{table*}
Besides the overall improved diagnosis across all 20 faults, our method presents superior performance in identifying those difficult-to-diagnose faults. For instance, for Fault 3, the F1-score improves from 82.0\% (GraphSAGE) to 97.0\% (our method). Similarly, for Fault 9 that is well-known for its low detection rates across many methods \cite{kovalenko2024graph}, the F1-score rises from 56.1\% to 71.9\%. Fault 15, another challenging case, sees an improvement from 61.0\% to 79.0\%. Fig.~\ref{fig10} shows the confusion matrix of the fault diagnosis results from our method based on the test data. Fig. \ref{fig12} visualizes the t-SNE plots of data samples in the 2D subspace before passing through the classifier, between the GraphSAGE and our method. It can be seen that with our model, more distinctive features are extracted for different fault classes. This explains the reason that our method can achieve superior fault diagnosis performance than GraphSAGE by the integration of hierarchical pooling, temporal modeling via LSTM, and global feature extraction (see Table~\ref{tab:fdr_score}).

\begin{figure*}[h]
	\centering
	\includegraphics[width=0.90\textwidth]{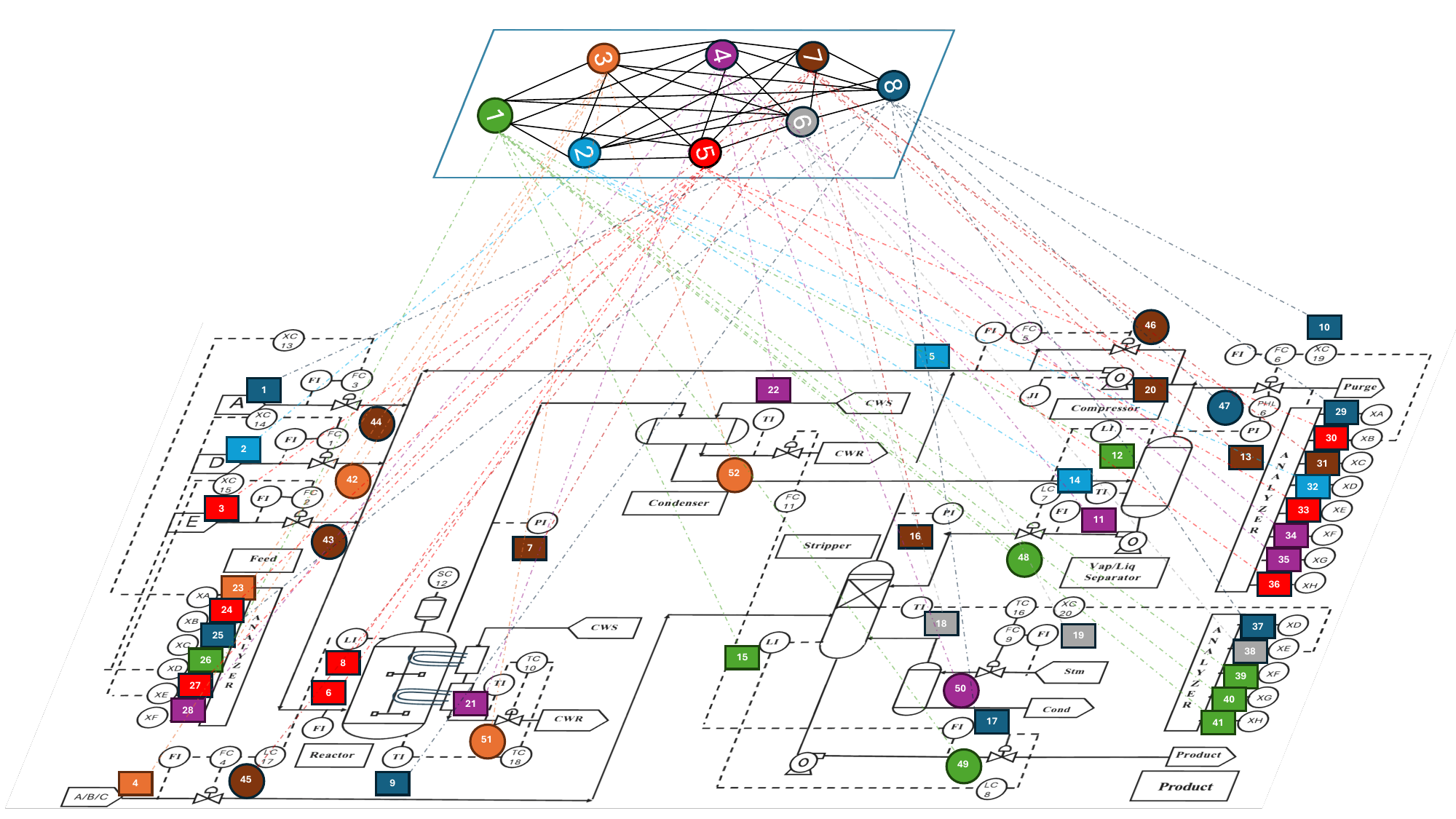}
	\caption{Illustration of multi-level pooling of nodes (sensors) into super-nodes for the TEP.}
	\label{fig13}
\end{figure*}

Fig. \ref{fig13} illustrates the multi-level clustering of process variables (sensors) from the TEP into super-nodes, where each cluster is visualized using a distinct color (super-nodes). It can be seen that, except for some variables, the multi-level pooling mechanism has grouped most of the sensors with strong links without the need for domain knowledge. For instance, sensors 15, 48, and 49 relating to the stripper are clustered together, which justifies representing these groups as super-nodes in the hierarchy. This provides another level of insights into the complex systems for improved fault diagnosis.

%%%%%%%%%%%%%%%%%%%%%%%%%%%%%%%%%%%%%%%%%%
\section{Conclusion}                                                                        
\label{sec: conclusion}
In this work, we proposed a graph-based fault diagnosis model that combines multi-level graph representation, temporal sequence modeling, and global feature extraction. The framework (LGF-MLTG) effectively models both local and global dependencies in multivariate time-series data by constructing dynamic spatio-temporal graphs. The experimental results show the superior performance of the proposed approach over existing baseline models. Detailed analysis were performed to validate the effectiveness of combining hierarchical graph structures with both temporal and global feature representations to capture complex dependencies. Future research may focus on extending this framework to incorporate causal discovery mechanisms to learn causal graph structures, and developing lightweight variants of the model for real-time deployment in edge devices.

\section*{Acknowledgments}
The authors acknowledge the support from Texas Tech University (TTU). B. Aryal acknowledges the Distinguished Graduate Student Assistantship program at TTU. B. Aryal also acknowledges Jose Powel for assisting with the figures.

\section*{Declarations}
\paragraph*{Funding} The authors acknowledge the support from Texas Tech University.
\paragraph*{Conflict of Interest} No potential conflict of interest was reported by the authors.
\paragraph*{Ethics approval} Not applicable.
\paragraph*{Consent for publication} Yes.
\paragraph*{Data availability} Not applicable.
\paragraph*{Code availability} Code will be made available on request.
\paragraph*{Author Contributions}
Bibek Aryal: Conceptualization, Methodology, Original draft, Data curation, Software, Visualization;
Gift Modweke: Validation, Writing - review \& editing;
Quigang Lu: Conceptualization, Methodology, Validation, Writing - review \& editing, Supervision.

%\bibliographystyle{IEEEtran}
%\bibliography{References}

\bibliographystyle{unsrt}
\bibliography{References}

\end{document}